# Design and Control of A Hybrid Sailboat for Enhanced Tacking Maneuver*

Ziran Zhang, Yiwei Lyu, Fahad Raza and Huihuan Qian

*Abstract*—Sailing robots provide a low-cost solution to conduct the ocean missions such as marine exploration, pollution detection, and border surveillance, etc. However, compared with other propeller-driven surface vessels, sailboat suffers in complex marine wind field due to its low mobility. Especially in tacking, sailboats are required to head upwind, and need to make a zig-zag path. In this trajectory, a series of turnings, which will cross the challenging no-go zone, place significant challenge as it will reduce speed greatly and consequently result in unsuccessful turning. This paper presents a hybrid sailboat design to solve this issue. Electric propellers and control system are added to a model sailboat. We have further designed the control strategy and tuned the parameters (PWM-time) experimentally. Finally, the system and control can complete the tacking maneuver with average speed approximately 10% higher and enhanced success rate, though the sailboat weight is much heavier.

## I. INTRODUCTION

Sailing robots have become a popular research topic as it provides a promising tool for long-range ocean surface exploration utilizing nature energy sources, e.g. wind, solar, wave, etc. They are also useful for data collection in hazardous or inhabitable environments, and coast surveillance or aiding, etc. [1]

There have been a lot of research efforts in the field of autonomous sailboats, including control, navigation, motion planning, and so on. The Microtransat Challenge [2] has been launched from 2010 with the aim to cross Atlantic Ocean with autonomous sailboats. A number of participating teams include Pinta [3], Breizh Spirit [4], Snoopy Sloop, and ABoat Time [5], etc. Agile motion control is one of the challenging tasks, as it provides mobility for demanding control, e.g. tacking in upwind paths, escaping from obstacles and beaches.

However, due to the relatively low mobility compared with other unmanned surface vehicles, sailboat should carefully examine complex wind field. On one hand, it should fully utilize the wind for propulsion. In upwind scenario, a zig-zag path should be used for tacking. On the other hand, when the heading angle is changed between the two close-hauled zones (Fig. 1), which also means that "No-go zone" has to be crossed, tacking performance will be poor, including low tacking successful rate and tacking efficiency. Sailboats usually lose speed when they approach to the sailboat turning point. In Fig. 2, when sailboat is in the turning point, it can hardly change position, while the heading angles (black arrow) changes in the "No-go zone".

To improve the tacking performance of sailboats, many studies have been conducted.

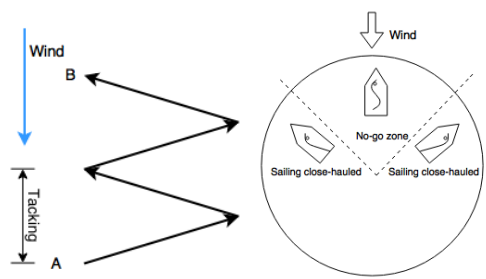

Figure 1. Basic concepts of sailing (*Tacking, "no-go zone", and close-hauled*)

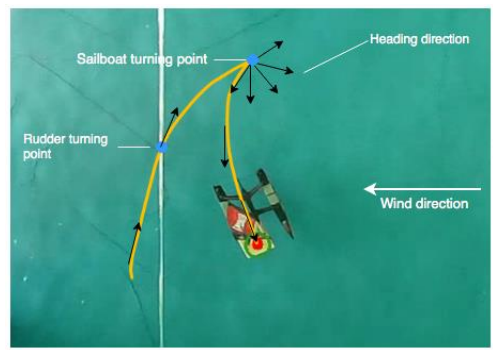

Figure 2. Typical tacking trajectory for traditional sailboat

Yang, Xiao and Jouffroy proved the existence of "no-go zone" through constructing "lift/drag" model and compared it with the "dead zone" model which was proposed earlier [6]. They eventually found that the two models shared similar characteristics [7], [8].

Jouffroy introduced a condition on the parametrization of the path for avoiding the vehicle to get stuck in the "no-go zone" [9], when he proposed a simple controller to steer the maneuvers on paths [10], [11]. A boundary demand for initial velocity was set as a condition, therefore, the control system he proposed was not completely controllable.

Saoud and other researchers conducted researches on sail angle optimization for an autonomous sailboat in sailing hoping to avoid the sailboat from entering the "no-go zone" to keep safe sailing, but the safety criterion established here was likely to reduce the tacking efficiency [12], [13], [14].

However, to fundamentally improve the tacking performance, another propelling force is a more feasible

* This paper is partially supported by supported by Project U1613226 supported by NSFC, the State Joint Engineering Lab and Shenzhen Engineering Lab on Robotics and Intelligent Manufacturing, Shenzhen, China, and PF.01.000143.

Ziran Zhang, Yiwei Lyu, Fahad Raza and Huihuan Qian are with The Chinese University of Hong Kong, Shenzhen. Corresponding author is Huihuan Qian, hhqian@cuhk.edu.cn

approach [15]. Nuno A. Cruz added a wing-sail to electric catamaran. Further study on how to coordinate the sail, rudder and electric propeller shall be conducted.

This paper, in another way, added two electric propellers to a sailboat, and examines the approach to coordinate sails, rudders and electric propellers. The advantages and limits brought by this hybrid power system are evaluated. It is validated that the hybrid system enhances the tacking success rate and turning efficiency significantly.

The paper is organized as follows. Section II introduces mechatronic design of the sailboat. Section III elaborates the strategy of motion control. In Section IV, comparison experiments are carried out, to examine the effect of additional mass and additional hydraulic drag. Data is also collected experimentally in this section. Analysis based on the data, are explicitly conducted in Section V. Section VI concludes the paper.

## II. MECHATRONIC DESIGN

Fig. 3 illustrates the overall component connection.

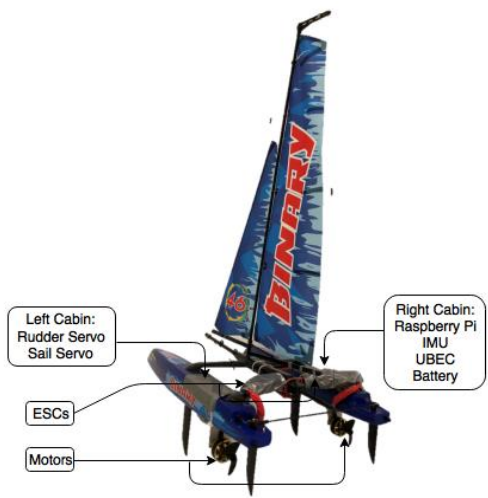

Figure 3. Assembly layout

A low-cost RF model catamaran sailboat (Binary 8807) is retrofitted to use in experiments, as it is more stable than a mono-hull sailboat. It consists of two servo motors (one each for sail and rudder control), one RF receiver, and 3 AAA batteries. We only use the two servo motors, and replace the other parts.

Two JFRC-U2305 brushless propellers are used to provide assistance during tacking. They are waterproof, and can be installed beneath the two hulls directly. Electronic Speed Controllers (ESC) are used to control and power the BLDC propellers.

A DC battery (Dualsky 3s-800mah-25c) is used as the power source in the boat. The rated voltage of the battery is 7.4V, and then converted to 5V through UBEC (fpv-3A) voltage transformer.

Raspberry Pi Zero is used as the microcontroller unit. It is small enough to be placed inside the cabin of the boat. An Inertial Measurement Unit (IMU) is selected to acquire orientation of the boat. Raspberry Pi communicates with a laptop (Laptop 2 in Fig. 4) through a bi-directional Wi-Fi.

Raspberry Pi sends out on-board information, e.g. boat orientation, to the laptop, while the laptop2 transmits control commands, e.g. sail and rudder angles and PWM for each propeller, to the boat.

The signal flow chart is illustrated in Fig. 4.

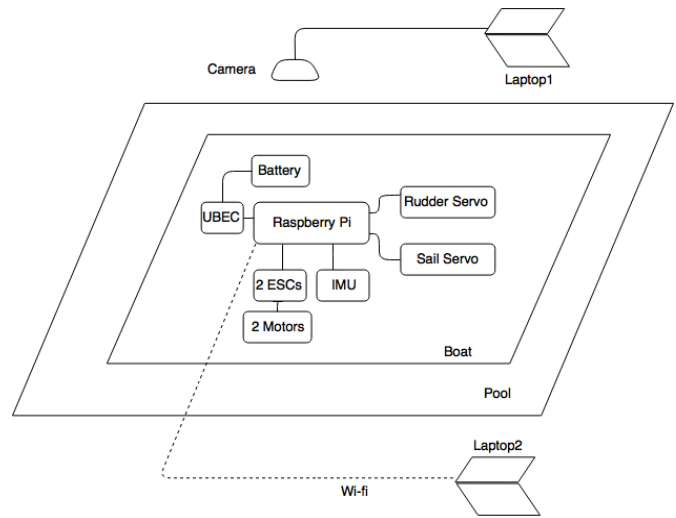

Figure 4. Details of experiment control part and data recording part

An overhang camera observes the water pool scenario, and is connected to laptop 1 for data recording. Code is written in Python 3.6 with OpenCV library. We utilize color marking board (a yellow board fixed to the surface of the sailboat) with object detection algorithm to boat localization.

In another part, i.e. the experiment control part, laptop2 uses Putty to connect the Raspberry Pi through Wi-Fi. The Raspberry Pi runs Python code to control the dynamic system and IMU. Pigio library is used to control propellers, and Adafruit-BN055 library is used to receive the orientation data.

To synchronize data from the two parts, timestamps are added to each data frame.

## III. CONTROL STRATEGY

Wind and electric propulsions are two driving forces for the hybrid boat. With solely electric energy, the motion control can work well. However, it can't last for long, as the electric energy are consumed at a very high rate.

The most crucial propelling assistance needed for sailboat should be offered at the turning point to go through the "No-go zone", when the propelling force on sail is minimal.

Fig. 5 shows the control strategy for tacking with sail and electric propeller. As tacking process starts, rudders rotate firstly to make a turn, and the propeller at the external side (farther away from the turning center) turns on to provide an auxiliary driving force, which helps sailboat get out of the "no-go zone" faster. After a successful turning, the propeller is turned off and the sailboat starts sailing normally with wind power only in close-hauled direction of the opposite side.

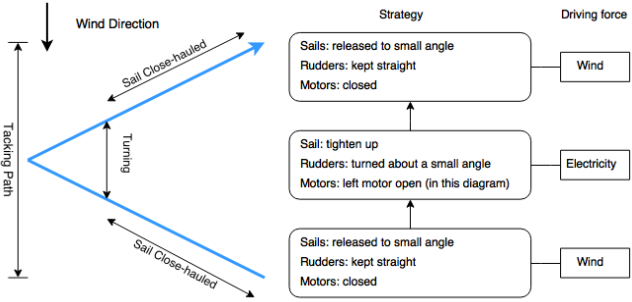

Figure 5. The control strategy of hybrid sailboat when tacking

## IV. EXPERIMENT FOR EFFECT WITH ADDED MASS

The inclusion of the electric propelling system adds extra mass to the boat. On one hand, it may increase inertia, which may increase the momentum to make taking easier. However, on the other hand, it decreases acceleration, and reduces speed. Moreover, the propeller in water will increase hydraulic drag and exert a negative effect in speed.

Due to the complex model in sailboat, we can hardly compare the dynamics of the sailboat before and after retrofitting analytically. Hence, in this section, we examine how much effect the additional mass has on the boat based on experiment. We set up a water pool (6m×10m) with controlled wind field (1.2-1.4m/s).

Non-retrofitted sailboat, heavier non-retrofitted sailboat (with same weight as the hybrid boat), and hybrid sailboat, are evaluated in tacking maneuvers (Fig. 6). The wind is blown from right to left by a line of electric fans. Each boat starts at the same point (1. Starting point) in the same angle of about 60° (2. Starting angle of attack). The rudder is in the neutral position, so that the motion is close to a line. When the boat achieved a certain displacement in x-direction, it starts turning by rudder or electric propeller. When it arrives at the position with the same y-coordinate, the tacking is completed, and the x-displacement can be named as the taking distance.

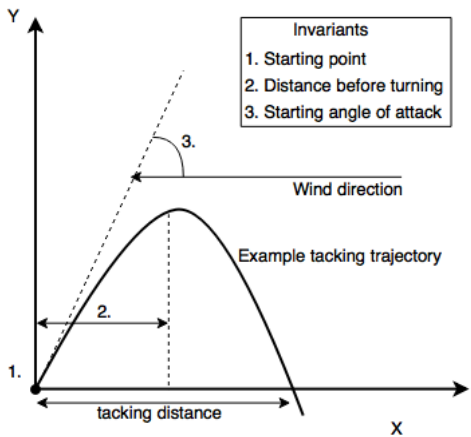

Figure 6. The experiment details about the constrains and variables

Table I illustrates the mass of the original none-retrofitted boat and the hybrid boat. Almost 67% has been increased in weight.

TABLE I. MASS OF SAILBOATS

| Object | Mass (g) |
|---|---|
| Modified sailboat | 691 |
| Non-retrofitted sailboat | 414 |
| Additional mass | 277 |

In the heavier non-retrofitted sailboat, we added 277g, evenly distributed to the two hulls. We manually controlled the sail and rudder for the three boats. In hybrid sailboat, the propellers are turned off. Table II sums up the experimental result. The tacking successful rates of the non-retrofitted sailboat, heavier non-retrofitted sailboat and hybrid sailboat are 10/35 (28.57% in 35 trials), 10/48 (20.83% in 48 trials) and 10/52 (19.23% in 52 trials), respectively. The distance is calculated in pixels, as can be seen in Fig. 8 from the overhang camera.

TABLE II. THE SUCCESSFUL TACKING RESULTS OF SAILBOATS

| Object | Avg. Tacking Distance (Pixel) | Avg. Tacking Time (s) | Avg. Tacking Speed (pixel/s) |
|---|---|---|---|
| Non-retrofitted sailboat | 176.0 | 18.2 | 9.67 |
| Heavier non-retrofitted sailboat | 61.1 | 17.8 | 3.43 |
| Hybrid sailboat (propellers off) | 61.5 | 19.8 | 3.11 |

It can be observed that with increased weight, the two heavier boats have worse performance in tacking, i.e. lower success rate and shorter tacking distance in x direction. Although their tacking time is similar, the significantly shorter tacking distance reduced the average tacking speed by around 65-68%.

However, compared with the heavier non-retrofitted sailboat, the underwater propellers do not significantly lower the tacking performance, and reduced the average tacking speed by only approximately 9%. Through this experiment, we conclude that in this size of model sailboat, increased inertia in weight deteriorate the mobility, and continue to seek for better control law to improve the performance in the next section.

## V. EXPERIMENT FOR PWM-TIME CONTROL LAW

PWM can be adopted as speed control for propellers. However, to cross the non-go zone, the heading angle should be controlled appropriately to close-hauled direction. Otherwise, it will either result in over-turning with lower efficient tacking, or under-turning into non-go zone.

In the turning point of tacking path, the external propeller is turned on with PWM, the internal propeller kept off. If the internal propeller is also on, but at a lower speed, turning is feasible, but the radius will be larger, beyond the size of the water pool and the view of the camera. Moreover, in this way, additional electric energy will be wasted for propelling, rather than turning. Experimentally, we study the tacking performance of various PWM-time.

### A. Various PWM-Time Experiments

We choose a number of PWM values (11%-21%, at the step of 2%). For each PWM, the hybrid sailboat is controlled manually 12 times, with different trials of time durations when PWM is on. Table III illustrates the turning results.

TABLE III. TURNING RESULTS FOR VARIOUS PWM VALUES

| PWM (%) | Time for Turning (s) | | | | | | | | | | | Average Time for Turning(s) |
|---|---|---|---|---|---|---|---|---|---|---|---|---|
| 11 | 3.75 | 3.70 | 3.24 | 3.87 | 3.98 | 3.76 | 3.69 | 3.65 | 3.95 | 4.35 | 4.51 | 4.04 | 3.93 |
| 13 | 3.05 | 3.29 | 3.04 | 2.98 | 3.13 | 2.84 | 3.19 | 2.87 | 2.91 | 3.48 | 4.36 | 2.96 | 3.07 |
| 15 | 2.48 | 2.85 | 2.74 | 2.56 | 3.31 | 2.38 | 2.33 | 2.96 | 2.75 | 2.25 | 2.10 | 2.43 | 2.45 |
| 17 | 1.97 | 2.09 | 2.26 | 2.61 | 2.15 | 1.77 | 2.09 | 2.22 | 2.16 | 1.96 | 2.12 | 1.60 | 2.11 |
| 19 | 2.08 | 2.32 | 1.84 | 1.48 | 1.67 | 1.67 | 1.74 | 1.69 | 1.80 | 2.60 | 1.80 | 2.03 | 1.71 |
| 21 | 1.72 | 2.08 | 1.57 | 2.03 | 1.23 | 1.34 | 1.45 | 1.62 | 1.32 | 1.42 | 1.45 | 2.45 | 1.40 |

The heads angle change is 120°, which is obtained from the testing video records. In this experiment, we measure the turning time, rather than the complete tacking time. This is because the time cost on sailing close-hauled doesn't change thus the turning time depends the tacking time. As shown in Fig. 7, turning (heading angle change) is the crucial part in tacking. If an appropriate heading angle is achieved after turning, the sailboat can accelerate rapidly.

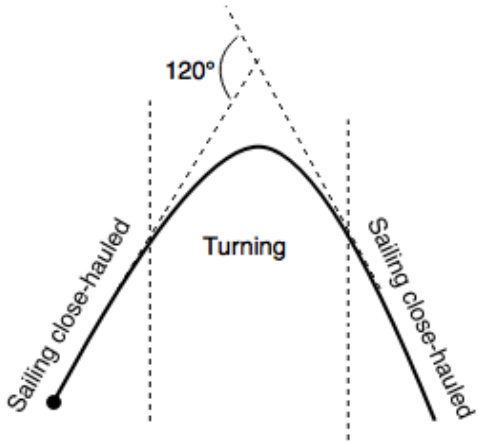

Figure 7. Path segment of tacking by hybrid sailboat

By means of a successful turning, the final heading angle is within the ±10° error region of the desired one. Data crossed by backslashes in TABLE III stands for unsuccessful results which is out of that error region.

The lower PWM value is, the slower the boat turns, and thus the easier we can manually tune the PWM time period. Hence, the success rate is higher when PWM value is low. However, the average time period is longer.

Turning efficiency and manual control complexity are two selection principles of PWM value. As a trade-off, 17% is selected for its higher turning efficiency (2.11 s) and successful rate (100%).

B. *Specific PWM –Time Experiment*

Thereafter, hybrid sailboat tacking experiments are conducted with PWM equals to 17%. To better illustrate how the control strategy works, Fig. 8 provides a real example of modified sailboat tacking, presenting the coordinate control operations among rudders, sails and propellers with corresponding working status.

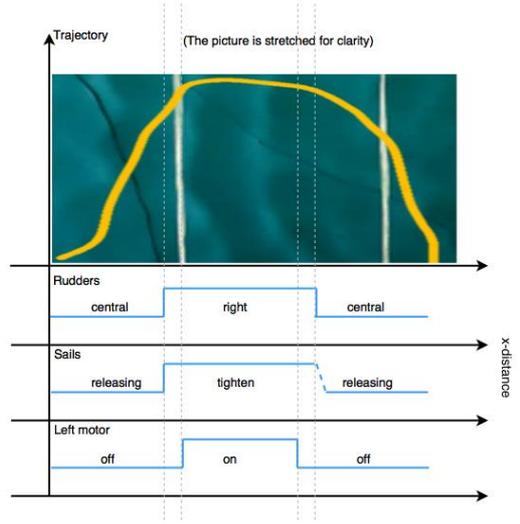

Figure 8. Control strategy used in a typical tacking trajectory

When the sails are released from tight status, it is marked by dot line, because sails' reaction to wind is not instantaneous, which takes a short duration (several 100ms). To provide a steering torque, reduce the energy consumption and increase the efficiency, rudders are turned first before the propeller is switched on.

Tacking results for modified sailboat are listed in TABLE IV, and the success rate is 100% (8 successful /8 trials). It can be seen that, with appropriate PWM values and time period, hybrid sailboat can have enhanced tacking performance significantly.

TABLE IV. THE SUCCESSFUL TACKING RESULTS OF MODIFIED SAILBOAT

| Trial # | PWM effective time | Tacking Distance (Pixels) | Tacking Time (s) | Tacking Speed (pixel/s) |
|---|---|---|---|---|
| 1 | 1.86 | 121 | 16 | 7.56 |
| 2 | 2.61 | 140 | 10 | 14.00 |
| 3 | 2.47 | 111 | 9 | 12.33 |
| 4 | 2.52 | 140 | 11 | 12.73 |
| 5 | 1.86 | 82 | 10 | 8.2 |
| 6 | 1.90 | 136 | 15 | 9.01 |
| 7 | 2.22 | 162 | 14 | 11.57 |
| 8 | 2.08 | 119 | 12 | 9.91 |
| Avg | 2.19 | 126 | 12 | 10.67 |

In the trial #2, 17% of PWM has been effective for 2.61 seconds. But the hybrid sailboat can still tackle successfully, unlike the 4th trial in Table III (PWM=17%). Some randomness exists in factors, e.g. initial heading, wind, wave, etc.

Fig. 9 shows three typical tacking trajectories of non-retrofitted sailboat (yellow line), heavier non-retrofitted sailboat (white line), and hybrid sailboat (red line). Turning command occurs when the boat arrives at the left vertical line (at the cross point). Before turning, the 3 tacking trajectories are similar. Due to the inadequate turning torque, non-retrofitted sailboats (both light and heavier version) keep forward motion along the similar trajectories, and achieve a stopping-and-turning. The heavier non-retrofitted sailboat is struggles in heading control, and is even blown back and consequently achieves shortest distance in tacking.

Hybrid sailboat with appropriate PWM control, has shorter y displacement, and can make sharper turn. After the heading angle is appropriately adjusted, the propeller is turned off, and sails keeps accelerating the boat.

The average tacking distances of non-retrofitted, heavier non-retrofitted, and hybrid sailboat are 176.0, 61.1 and 126.0 pixels respectively from TABLE II&IV.

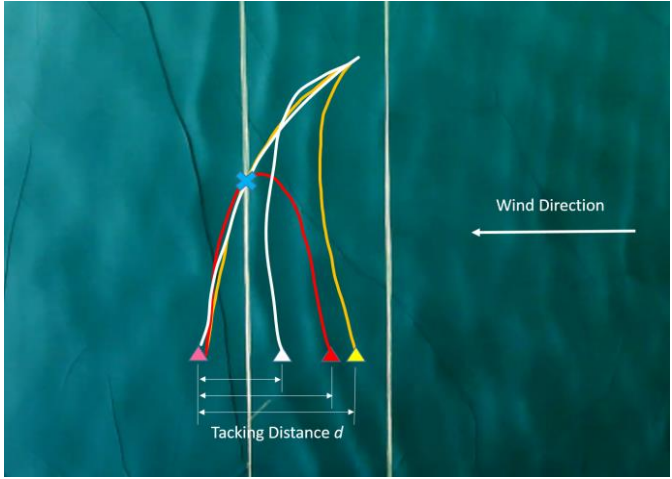

Figure 9. Typical tacking trajectories of original boat (yellow line), original sailboat with additional mass (white line) and modified sailboat (red line).

For one tacking maneuver, hybrid sailboat is shorter than the non-retrofitted sailboat. However, as the duration to complete the tacking is about 34% shorter, the average speed in tacking direction is approximately 10% higher.

### C. PWM –Time Curve Fitting

A scatter plot corresponding to TABLE III is provided in Fig. 10, which provides different PWM values and their corresponding turning time. Circle stands for successful turnings while cross represents unsuccessful turnings.

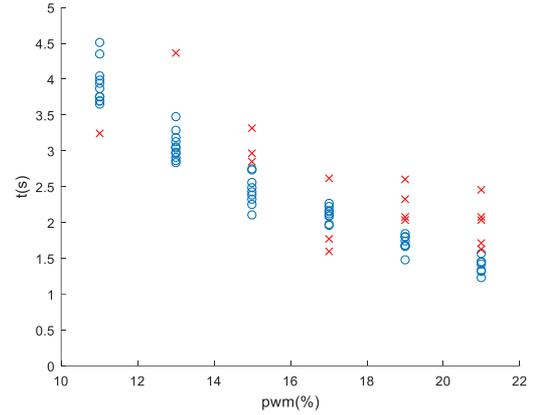

Figure 10. The scatter plot of PWM values and their turning time

Error data are eliminated and potential rules are observed in Fig 10. Various order polynomial curve fittings are conducted on the data (Table V), and it is found that the cubic function model has the best fitting with least RMSE (root-mean-square error).

TABLE V. CURVE FITTING RESULT

| Fit type | Poly1 | Poly2 | Poly3 | Poly4 | Poly5 |
|---|---|---|---|---|---|
| RMSE | 0.23 | 0.18 | 0.17 | 0.18 | 0.18 |

By the dots plotting in Fig. 11, the range of each dot set narrows down when PWM value increases, indicating that the manual control successful rate decreases when PWM value increases. As the time-period for turning is shorter, it is not easy to turn it manually.

For the fitting curve, the effective turning time can be tried for a certain higher PWM. It will be useful in future tests to find out the right turning period.

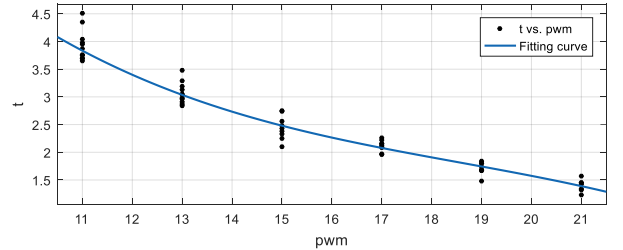

Figure 11. Fitting curve for PWM values to their turning time

### VI. CONCLUSION AND FUTURE WORK

This paper attempts to address the problem of turning inefficiency and low successful rate in conventional sailboats tacking process, with a solution of hybrid sailboat system with both sail and electric propellers, as well as control strategy which involves coordinated control of sails, rudders and propellers. Validity of the hybrid power system is examined with clear evidence of improvement in turning efficiency and successful tacking rate.

A method on how to choose the proper PWM value is developed, and PWM-time relationship is proposed which can be helpful for future work as reference.

Further researches can be expected on mass reduction of the modified sailboat and robust control for autonomous navigation. It is believed that modified sailboat with mass reduction will have more efficient tacking performance.